\pdfoutput=1

\documentclass[11pt]{article}

\usepackage[final]{acl}

\usepackage{times}
\usepackage{latexsym}

\usepackage{pbox}
\usepackage{makecell}
\usepackage{graphicx}
\usepackage{subcaption}
\usepackage{setspace}
\usepackage{enumitem}
\usepackage{booktabs, cellspace, multirow}
\usepackage{amsmath}
\usepackage{amssymb}
\usepackage{soul}
\usepackage{xspace}
\usepackage{caption}
\usepackage{tabularx}
\usepackage{verbatim}

\newcommand\Tstrut{\rule{0pt}{2.6ex}}         
\newcommand\Bstrut{\rule[-1.2ex]{0pt}{0pt}}
\usepackage[T1]{fontenc}

\usepackage[utf8]{inputenc}

\usepackage{microtype}

\newcommand{\capdataname}{DialFact}
\newcommand{\dataname}{\textsc{DialFact}\xspace}
\newcommand{\modelname}{Aug-WoW\xspace}
\newcommand{\dataurl}{https://github.com/salesforce/DialFact}

\title{{\capdataname}: A Benchmark for Fact-Checking in Dialogue}

\author{Prakhar Gupta$^\dag$, Chien-Sheng Wu$^\ddag$, Wenhao Liu$^\ddag$, Caiming Xiong$^\ddag$ \\
  Language Technologies Institute, Carnegie Mellon University$^\dag$ \\
  Salesforce AI Research$^\ddag$ \\
  \texttt{prakharg@cmu.edu, \{wu.jason,wenhao.liu,cxiong\}@salesforce.com} \\}

\begin{document}
\maketitle

\begin{abstract}
Fact-checking is an essential tool to mitigate the spread of misinformation and disinformation. We introduce the task of fact-checking in dialogue, which is a relatively unexplored area.
We construct {\dataname}, a testing benchmark dataset of 22,245 annotated conversational claims, paired with pieces of evidence from Wikipedia.
There are three sub-tasks in {\dataname}: 1) Verifiable claim detection task distinguishes whether a response carries verifiable factual information; 2) Evidence retrieval task retrieves the most relevant Wikipedia snippets as evidence; 3) Claim verification task predicts a dialogue response to be supported, refuted, or not enough information.
We found that existing fact-checking models trained on non-dialogue data like FEVER~\cite{thorne-etal-2018-fever} fail to perform well on our task, and thus, we  propose  a  simple  yet  data-efficient solution to effectively improve fact-checking performance in dialogue.
We point out unique challenges in {\dataname} such as handling the colloquialisms, coreferences and retrieval ambiguities in the error analysis to shed light on future research in this direction\footnote{Data and code are available at \url{\dataurl}}.

\end{abstract}

\section{Introduction}
\label{sec:introduction}

Misinformation online can have deleterious consequences to our society, especially during public health crises like the COVID-19 pandemic.
False and outdated information can be spread not only by humans but also by automatic agents as generative models have shown remarkable progress recently~\cite{adiwardana2020towards, xu2021beyond}. 
These systems are not perfect, as they can either generate hallucinated and imperfect information, or they can be abused to automatically generate false claims and spread misinformation at a massive scale. 
Fact verification tools are thus necessary in the current information age to tackle the spread of misinformation propagated.

Fact-checking was introduced in~\citet{wang-2017-liar,thorne-etal-2018-fever} and since then a growing body of research has explored and suggested various tasks and resources to address the challenges in this area.
Fact-checking has been explored in medium such as 
Wikipedia passages, tables, social media and news articles~\cite{guo2021survey,10.1145/3485127}.
In dialogue domain, related work either focus on evaluating factual consistency~\citep{honovich2021q, qin-etal-2021-dont} or consistent response generation~\citep{rashkin-etal-2021-increasing,shuster2021retrieval}. 
However, due to lack of publicly available benchmarks, fact checking is still underexplored in the dialogue domain.

\begin{table}[tb]
\small
    \centering
    \begin{tabularx}{\columnwidth}{l}
\toprule
    \textbf{Dialogue Context:} I have family in Ireland!   Have you \\ ever been there? \\
    \textbf{Evidence: } Ireland is an island in the North Atlantic.\\
    \midrule
    \textbf{Non-Verifiable Response:} I haven't been but want to! \\
    \textbf{Verifiable Supported Response:} I haven't. It is \\ an island in the north Atlantic right? \\
    \textbf{Verifiable Refuted Response:} I haven't been. Isn't it \\somewhere in north Pacific? \\
    \textbf{Verifiable NEI Response:} I haven't been. I heard it's \\the most popular tourist location in Europe! \\
    
\bottomrule
    \end{tabularx}
    \captionof{figure}{Dialogue fact-checking involves predicting if a response should be considered a Verifiable claim, followed by finding relevant evidence, and finally predicting if the it is \textsc{Supported, Refuted} or \textsc{NEI}.}
    \label{tab:teaser}
    \vspace{-1em}
\end{table}

Verifying factual correctness of claims in dialogue poses new challenges to both dataset construction and modeling. Claims in existing datasets are from formal sources such as news articles and they are generally succinct and formal. In contrast, claims in dialogue are often informal and sparse in factual content. Furthermore, dialogue utterances often include personal opinions, slang, and colloquialisms which need to be distinguished from factual information. 
Another challenge in dialogue fact-checking is that ellipsis and coreference occur frequently which make utterances incomplete and ambiguous~\cite{devault-stone-2007-managing}. 
Although humans can easily understand utterances with references or absent information based on the dialogue context and their reasoning skills, a fact-checking system may need to model this behavior explicitly. 

We introduce the task of fact-checking in dialogue and propose an evaluation dataset, {{\dataname}}. 
An example is shown in Figure~\ref{tab:teaser}. 
{\dataname} has three sub-tasks: 1) Verifiable claim detection aims to distinguish responses that do not contain verifiable factual information, such as ``I haven't been but want to!'' in Figure~\ref{tab:teaser}. 2) Evidence retrieval involves selecting the most relevant knowledge snippets from Wikipedia which can verify the response. 3) Claim verification aims to classify if a response is supported, refuted, or does not have enough information to verify the response given the dialogue history and the retrieved evidence. 

{\dataname} consists of both human-written and machine-generated claims based on the Wizard of Wikipedia~\cite{dinanRSFAW19} dialogue dataset. Each response claim and its evidence sentences from Wikipedia are annotated by crowd workers and we perform rigorous quality checks on the annotations. For fact verification, we propose creation of weakly-supervised training data by leveraging techniques such as negation, entity swapping, language model mask-and-fill, and knowledge-grounded generation. We establish baseline model performance on this task, and point out the weaknesses of fact-checking models. Our analysis show that this is a non-trivial task with challenges remaining for future work. We hope that future work can leverage this dataset as a fact-checking benchmark or for development of automatic consistency metrics, and advance the state-of-the art in knowledge-grounded dialogue generation and evaluation.

\section{Related Work}
\label{sec:related}
\vspace{-2mm}
\noindent
\textbf{Fact Verification}
The spread of false information online has led to a growing body of research exploring automatic fact-checking.
\citet{thorne-etal-2018-fever} and subsequent works~\cite{2019TabFactA,jiang-etal-2020-hover, norregaard-derczynski-2021-danfever, aly2021feverous} introduced fact extraction and verification datasets verifiable against pieces of evidence from Wikipedia articles. 
Fact-checking has been explored in a variety of mediums such as 
Wikipedia based claims~\cite{schuster-etal-2021-get}, 
claims over tables~\cite{aly2021feverous}, 
scientific claims~\cite{wadden-etal-2020-fact}, 
and social media claims~\cite{10.1007/978-3-030-72240-1_75}.
However, fact-checking in dialogue is still an underexplored area.
\citet{kim-etal-2021-robust} explored fact-checking for colloquial claims, curated by converting FEVER claims into colloquial style. Although closely related to our work, colloquial claims is not a dialogue dataset, only contains verifiable claims, and does not have dialogue contexts for claims. In \dataname, on the other hand, both evidence retrieval and claim verification are more challenging as they require resolving ambiguities and coreferences from the dialogue context.

\vspace{5pt}
\noindent
\textbf{Consistency in Dialogue}
Neural dialogue systems grounded on knowledge sources such as Wikipedia~\cite{dinanRSFAW19}, knowledge graphs~\cite{wu-etal-2019-proactive} or snippets from the internet~\cite{komeili2021internet} have garnered interest in recent years. 
Despite generating plausible and engaging responses, existing models still hallucinate invalid information \cite{roller-etal-2021-recipes}. Ensuring safety and consistency in dialogue response generation is thus an actively explored area~\cite{rashkin-etal-2021-increasing, shuster2021retrieval}.
Some recent works have proposed evaluation metrics and benchmarks for factual consistency in knowledge grounded response generation~\cite{honovich2021q,dziri2021evaluating}. Our work instead focuses on fact-checking in dialogue for both human and machine-generated responses, and involves additional tasks of verifiable claim detection and evidence retrieval.

\vspace{5pt}
\noindent
\textbf{Synthetic datasets}
Synthetic dataset construction has been shown to improve robustness of evaluation models~\cite{gupta-etal-2021-synthesizing, ghazarian-etal-2021-plot} and improve the complexity of test sets~\cite{Sakaguchi10, feng-etal-2021-survey}. Synthetic claims have been explored in fact-checking to create hard test sets. Several participants in the FEVER 2.0 breakers phase~\cite{niewinski-etal-2019-gem, hidey-etal-2020-deseption, atanasova-etal-2020-generating} proposed approaches for automatically generated adversarial claims. Recently, \citet{jiang-etal-2020-hover} created complex multi-hop claims using word substitutions, \citet{saakyan-etal-2021-covid} used Bert based token-infilling to created refuted claims, and \citet{schuster-etal-2021-get} created synthetic revisions to Wikipedia sentences to improve fact-checking robustness. Our work also introduces techniques to create synthetic claims in the context of dialogue fact-checking.
\section{Task Background}
\label{sec:background}

Let a conversation context consist of a list of utterances $C= \{u_1,u_2,...,u_n\}$. The task is to perform fact-checking on the last utterance of the conversation $u_n$, henceforth called claim $c$. 
Fact-checking claims in conversations is a pipeline that consists of several steps. First, the system needs to decide whether a response is \textsc{Verifiable} or \textsc{Non-Verifiable}. We define them as follows:
\textbf{\textsc{Non-Verifiable:}}
The claim contains no verifiable factual information. It includes claims with personal opinions or personal information. 
\textbf{\textsc{Verifiable:}} The claim contains at least one factual information verifiable against a background corpus (Wikipedia in this task).

Next, the system should retrieve documents from the background corpus and select relevant evidence sentences from the documents. Finally, the system should predict whether the claim belongs to one of the following three categories: 
\textbf{\textsc{Supported:}}
The response contains factual information which is valid in light of the evidence.
\textbf{\textsc{Refuted:}}
The response contains factual information which is invalid in light of the evidence.
\textbf{\textsc{NotEnoughInformation (NEI):}}
The response contains factual information which can not be validated (supported or refuted) with the evidence. 

\textsc{Verifiable} claims can be \textsc{Supported}, \textsc{Refuted}, or \textsc{NEI}, and \textsc{Non-Verifiable} claims are always \textsc{NEI}. We leverage the \textit{Wizard of Wikipedia} (WoW) dataset~\cite{dinanRSFAW19} as the base to build this task. WoW is a knowledge-grounded open-domain dialogue dataset with conversations between two speakers - a wizard who has access to background Wikipedia documents to deliver knowledge carrying responses, and an apprentice who plays the role of a curious learner. 
For each turn $u_i$, the wizard is shown a set of articles $K_i$ retrieved from Wikipedia. The wizard either chooses a relevant knowledge sentence $k_i$ from the set $K_i$, or chooses a \textit{no sentence used} option to construct a response. 
For our fact-checking task, we additionally need claims which belong to \textsc{Refuted} and \textsc{NEI} categories. We next describe the methodologies used to create claims from the valid and test splits of the WoW dataset.

\noindent
\section{Dataset Construction and Annotation}
We use two approaches to create claim responses for \dataname: 1) Automatically generated claims, and 2) Human written claims to emulates claims created by dialogue systems and humans respectively. All claims are further annotated by crowd workers on Amazon Mechanical Turk (Mturk). 

\subsection{Automatically Generated Claims}
In this approach, we use automatic methods to create claims for all categories either from scratch or by mutating the responses in WoW dataset.

\subsubsection{Methods for claim generation}
\label{sec:augmentation}

\noindent
\textbf{Negation}
We use the 42 rule-based transformations from \citet{thorne-etal-2019-evaluating} which apply to verb phrases of the claims to convert them to their negated versions by adding words like ``not'' or ``no''. It typically creates \textsc{Refuted} claims.

\noindent
\textbf{Substitution}
We perform three types of substitutions:
For 1) Context and knowledge-based entity substitution,
we first run SpaCy NER tagging~\cite{spacy2} on a response $u_i$ from WoW. 
We then swap an entity in the response $u_i$ with an entity from either its conversation context $C$ or its background knowledge articles set $K_i$. An entity is only swapped if it is present in $k_i$, the original knowledge sentence to avoid swaps which do not change the facts. Entities are swapped within their types.
For 2) Sense-based substitution, we swap an entity in $u_i$ with an entity with a similar ``sense'' returned from the sense2vec~\cite{trask2015sense2vec} library. 
For 3) Adjective substitution, we substitute adjectives in a claim (ignoring adjectives related to emotions, such as ``happy'') with their WordNet~\cite{miller1998wordnet} antonyms (for example \textit{best} is replaced with \textit{worst}). These operations typically create \textsc{Refuted} claims.

\noindent
\textbf{Mask-and-Fill} This method generates claims in two stages: 1) Mask salient words from the original claims, and 2) Substitute those words with their alternates using a language model. For masking salient words in the original response claims, we follow the procedure from \citet{thorne-vlachos-2021-evidence} and use the Neutrality Masker model from \citet{shah2020automatic}. It predicts the tokens which upon masking are likely
to cause a label flip from \textsc{Supported} to \textsc{NEI}. 
For step 2) we first train a T5-base model~\cite{2020t5} on the WoW dataset on the task of infilling masked tokens conditioned on evidence sentences. For training, the input sequence consists of concatenated evidence sentence $k_i$, dialogue context $C$, and the gold response with masked spans at random positions, and the output is the gold response. The model is thus trained to infill a masked response based on the provided evidence and the dialogue context. For generating response claims which belong to \textsc{Refuted} or \textsc{NEI} categories, we use the following types of evidence sentences to condition the infilling: a) empty evidence, b) evidence sentences selected randomly from the knowledge article set $K_i$ belonging to the original response, and c) evidence sentences from a Wikipedia article of an entity retrieved using sense2vec based on its similarity with the entities in the original response. Conditioning on such evidence lead to generation of claims which have factual details inconsistent with the original evidence.

\noindent
\textbf{Generation}
We fine-tune one of the best chit-chat dialogue systems, Blenderbot model~\cite{roller-etal-2021-recipes}, on the WoW dataset. The model takes the concatenation of the knowledge sentence $k_i$ and the dialogue context $C$ as input and it is trained to predict the tokens of the gold response. To generate new response claims, we condition the model on the three types of evidence described in the Mask-and-Fill approach. We use a high temperature (1.5) and nucleus sampling~\cite{HoltzmanBDFC20} with $p=0.9$ during decoding to encourage the model to generate unexpected and non-contextual entities in the responses.

\noindent
\textbf{Final claim set creation} Our target is to create a challenging and diverse test set for dialogue fact-checking. 
Using the aforementioned methods of claim generation, we get a set $R_c=\{r_1,r_2,...,r_k\}$ of response claims for a dialogue context $C$. 
To select a final set of claims, we first remove any responses which do not have at least 3 words different from other responses in $R_c$, then filter out less fluent claims whose GPT-2~\cite{radford2019language} perplexity scores are higher than 1.1 times the average perplexity scores of the responses in $R_c$. 
We then score the response claims using existing state-of-the-art models related to our task: namely Dialogue NLI~\cite{welleck-etal-2019-dialogue}, Dialogue contradiction detection~\cite{nie-etal-2021-like}, FEVER based fact verification~\cite{schuster-etal-2021-get} and fact-checking on colloquial claims~\cite{kim-etal-2021-robust}. 
For each model, we calculate the entropy of the scores predicted for each label and rank the claims in $R_c$ based on the sum of the entropy of the scores of all the models, which gives an estimate of the confusion or difficulty in classifying the claims.
The top 4 responses from the ranked list are chosen as the final set of response claims for that context. 


\subsubsection{Evidence set creation}
\label{sec:evidence}
For each claim, a set of evidence sentences is first automatically created and then labelled by crowd workers. We first extract a set of named entities and noun phrases $n_k$ from the following sources: the claim $c$, the dialogue context $C$, the original response $u_i$ for the dialogue context in WoW, and the title of the knowledge articles $K_i$ shown to the wizard for $u_i$. We use the MediaWiki API\footnote{\label{note1}\url{www.mediawiki.org/wiki/API:Main_page}} to find a set of relevant Wikipedia pages $P_c$ for $n_k$. We then create a set of candidate sentences with the first 10 sentences of each page in $P_c$. Finally, we use two methods - SpaCy's word2vec similarity\footnote{\url{www.spacy.io/}} and BM25 similarity\footnote{\url{www.github.com/dorianbrown/rank_bm25}} to rank the top 10 evidence sentences using each method. We then combine the non-overlapping evidence from both methods to create the final evidence set $e_c$ for each claim $c$. We add the knowledge sentence $k_i$ associated with the original response in the WoW dataset if it is not already present in $e_c$.

\subsubsection{Claim and Evidence Annotation}
\label{sec:claimannotion}
We carry out the annotations of the claims and evidence on the Mturk platform in 3 rounds. The screenshot of the annotation UI is shown in Figure~\ref{fig:interface1} of the Appendix. In each round a worker sees the claim $c$, its dialogue context $C$, and its associated evidence sentences $e_c$. Workers have to perform 3 tasks: First, they select if the claim is \textsc{Verifiable} or \textsc{Non-Verifiable}. Second, they select one or more evidence sentences related to the response claim. In case the set of evidence shown is not enough to decide the label of the response, or if they choose NEI, they are instructed to search Wikipedia and add relevant additional evidence sentences in the interface. For NEI claims they are instructed to add evidence sentences which are most related to the claim. Third, they choose the category of the response - \textsc{Supported, Refuted}, or \textsc{NEI}. For \textsc{Non-Verifiable} claims, NEI is auto-selected. Since automatically created responses can have grammatical or coherence related issues, in the first round of labeling, annotators are asked to edit a response to make it appropriate to the context if needed, or mark a response as incoherent, in which case it is removed from further rounds (We dropped 5\% of incoherent claims). 
In the second and third rounds we gather 2 additional annotations for each claim. We select the label which has the majority vote among the set of 3 annotations across all rounds. The evidence set for each claim is the union of evidence annotated in any of the rounds.
Note that this mechanism can miss relevant evidence sometimes due to either retrieval errors in evidence set creation, or insufficient search of evidence or incorrect evidence annotation by workers.

\subsection{Human Written Claims}
Our dataset also consists of human written claims to cover lexical and stylistic patterns present in human-human conversations.
The annotation is carried out in 3 rounds. \textit{In the first round}, we instruct crowd workers to write \textsc{Verifiable} factual responses conditioned on dialogue context and a set of evidence sentences for a pre-specified label $l_c$ - one of \textsc{Supported, Refuted}, or \textsc{NEI}.
Workers were provided detailed examples and instructions for the task such as ``Avoid using negation words such as do not, no for Refuted claims'' (Appendix~\ref{sec:amt_info}).
The evidence set for each claim is constructed using the method described in section~\ref{sec:evidence}. 
\textit{In the second round}, we use the claim labeling interface from section~\ref{sec:claimannotion} to gather labels for the claims collected in the first round. For any claim which is not labeled in the second round with the original label $l_c$, we gather a third round of annotations. If the label in the third round does not match $l_c$, we drop that claim from the dataset. We drop about 7\% of the human written claims.

\begin{table}[t]
\small
\addtolength{\tabcolsep}{-3pt}
\begin{tabular}{lccccc}
\toprule
 & \multicolumn{4}{c}{\textbf{Validation}}                                                                                                         &       \\ \hline
\multicolumn{1}{l|}{}          & Supported & Refuted & \begin{tabular}[c]{@{}c@{}}NEI-\\ Factual\end{tabular} & \begin{tabular}[c]{@{}c@{}}NEI-\\ Personal\end{tabular} & Total \\ \hline
\multicolumn{1}{l|}{Generated} & 1686      & 1047    & 150                                                    & 1745                                                    & 4628  \\ 
\multicolumn{1}{l|}{Written}   & 1656      & 2316    & 1836                                                   & 0                                                       & 5808  \\ 
\multicolumn{1}{l|}{Total}     & 3342      & 3363    & 1986                                                   & 1745                                                   & 10436 \\ \hline 
\noalign{\vskip 1pt}  
 &
\multicolumn{4}{c}{\textbf{Test}}                                                                                                               &       \\ \hline
\multicolumn{1}{l|}{}          & Supported & Refuted & \begin{tabular}[c]{@{}c@{}}NEI-\\ Factual\end{tabular} & \begin{tabular}[c]{@{}c@{}}NEI-\\ Personal\end{tabular} & Total \\ \hline
\multicolumn{1}{l|}{Generated} & 2446      & 1195    & 1278                                                   & 1305                                                    & 6224  \\ 
\multicolumn{1}{l|}{Written}   & 1493      & 2740    & 1268                                                   & 84                                                       & 5585  \\ 
\multicolumn{1}{l|}{Total}     & 3939      & 3935    & 2546                                                   & 1389                                                    & 11809 \\ 
\bottomrule
\end{tabular}
    \caption{Dataset statistics of \dataname for all categories and splits. \textit{Generated} denotes automatically generated and \textit{Written} denotes human written claims.}
    \vspace{-1.0pc}
    \label{tab:statistics}
\end{table}

\begin{table*}[htb]
\small
\centering
\begin{tabular}{lcc|lcc|lcc}
\toprule
\multicolumn{3}{c|}{All}          & \multicolumn{3}{c|}{Labelled}       & \multicolumn{3}{c}{Written}   \\ 
\hline
Bigram            & LMI & p(l/w) & Bigram            & p(l/w) & p(l/w) & Bigram       & p(l/w) & p(l/w) \\ \hline
he was            & 396 & 0.45   & he was            & 692    & 0.40   & only one     & 201    & 0.95   \\
was born          & 362 & 0.64   & singer songwriter & 471    & 0.61   & referred as  & 169    & 0.83   \\
spectrum visible  & 195 & 0.80   & spectrum visible  & 447    & 0.82   & drama school & 163    & 0.89   \\
visible light     & 188 & 0.76   & visible light     & 431    & 0.74   & harry potter & 160    & 0.60   \\
on spectrum       & 186 & 0.73   & on spectrum       & 431    & 0.78   & pins are     & 158    & 0.83   \\
an american       & 177 & 0.50   & an american       & 391    & 0.47   & only be      & 152    & 0.89   \\
\bottomrule
\end{tabular}
\caption{Top bigrams in the test set for \textsc{Refute} category. \dataname does not include bias based on obvious negations such as ``do not'' and ``is not''. }
    \vspace{-1.0pc}
    \label{tab:lmi}
\end{table*}

\subsection{Dataset Statistics}
We present the dataset statistics in Table~\ref{tab:statistics}. The dataset consists of balanced \textsc{Supported} and \textsc{Refuted} claims. Test set contains claims for 3,760 dialogue contexts with an average of 3.1 claims per context, and validation contains claims for 3,738 contexts with an average of 2.8 claims per context. The average number of tokens per claim is 22.0 in test set and 20.0 in validation set. Average number of evidence per claim is 1.3 in the test set and 1.1 in the validation set. We show some sample instances in Table~\ref{tab:examples} in the Appendix.

\subsection{Quality Control}
\label{sec:quality}

\noindent
\textbf{Annotators}: We hire workers on Mturk with with at least 5000 HITS done and an acceptance rate of 95\% or above. Workers have to first pass a qualification test where they are shown the task instructions, label definitions, and multiple examples and the explanations for each label. Then they are asked to label or write 12 claims. Using these qualification tests, we get a final set of 87 workers for the main data collection stage (Appendix~\ref{sec:amt_info}). 

\noindent \textbf{Quality checks}
Annotations were carried out in batches over multiple weeks. We examined random samples to provide feedback to workers. Workers with poor annotations were either asked to retake a new qualification test or removed from further batches. We recollected annotations for data annotated by removed workers.
We provide tooltips and examples during annotation, and we also added automatic checks to alert workers about issues such as too short responses, no evidence selected, and copy-pasting evidence sentences as claims.

\noindent \textbf{Data validation}
To evaluate inter-annotator agreement, we collected 2 extra rounds of annotations for 1200 claims for both automatically generated and human written claims, which is 10\% of the data. 
Krippendorff's alpha value for category labels was 0.68 for human written claims and 0.58 for automatically generated claims, denoting moderate agreement.
Krippendorff's alpha for \textsc{Verifiable} versus \textsc{Non-Verifiable} was 0.49, with a low-to-moderate agreement. The lower agreement is due to some claims like ``Guns N' Roses was the greatest rock band of all time.'', where it is difficult to judge if this is a personal opinion or a verifiable fact. In such conflicts, workers would still typically correctly label such ambiguous claims as NEI. 

\noindent
\textbf{Lexical Biases}
Following~\citet{schuster-etal-2019-towards}, we measure the Local Mutual Information (LMI) to measure the correlation between bigrams in the claims ($w$) and the categories $l$, defined as follows:
$LMI(w,l)=p(w,l)log\left(\frac{p(l/w))}{p(l))} \right )$.
We present the top bigrams in \textsc{Refuted} claims and their LMI value in Table~\ref{tab:lmi}.
The top bigrams in \dataname do not include obvious negations such as ``do not'', ``is not'', are mostly topical in nature, and the $p(l/w)$ value is low with the Refute label. 
Investigating generated and written claims separately, we found that bigrams such as ``does not, only one, did not, are not'' had higher $p(l/w)$ in written claims compared to generated claims for \textsc{Refuted} category, although their LMI values were not high. Finally, there is significant overlap between the top bigrams for different categories, suggesting an absence of obvious lexical biases in the dataset.

\section{Experiments}
\label{sec:experiments}
We propose new baselines and compare with existing models for three sub-tasks in dialogue fact-checking - 1) Verifiable claim detection, 2) Evidence retrieval, and 3) Claim verification.

\begin{table}[t]
\centering
\addtolength{\tabcolsep}{-3pt}

\resizebox{\linewidth}{!}{
\begin{tabular}{lcccc}
\toprule
Baseline & Accuracy & Verifiable F1 & {\begin{tabular}[c]{@{}l@{}}Non-Verifiable F1\end{tabular}} \\ \hline
Random & 50.0  & 64.2 & 19.2\\
\begin{tabular}[c]{@{}c@{}}Lexical \end{tabular} & 79.4 &88.1 & 33.8\\ 
DNLI & 82.1   & 89.9  & 37.1   \\
\begin{tabular}[c]{@{}c@{}}Lexical+DNLI\end{tabular} & \textbf{82.8} & \textbf{90.2} & \textbf{39.1} \\
\bottomrule
\end{tabular}
}
    \caption{Accuracy and Macro F1 scores for {Verifiable} claim detection on the test set.}
    \vspace{-6mm}
    \label{tab:claimcarrying}
\end{table}

\subsection{Verifiable Claim Detection}
\label{sec:claimdetection}
We propose three simple baselines for verifiable claim detection. 
1) \textit{Lexical overlap} calculates the maximum word overlap between a claim and all evidence sentences after removing punctuation and stopwords using SpaCy. 
2) \textit{DNLI} uses the probability of the neutral class from the Dialogue Natural Language Inference model~\cite{welleck-etal-2019-dialogue}. 
3) \textit{Lexical+DNLI} uses the sum of scores of both baselines and \textit{Random} predicts each class with 50\% probability. For all baselines, we mark a response as \textsc{Verifiable} or \textsc{Non-Verifiable} based on a threshold value selected using validation data. We present the accuracy and individual F1 scores for both classes in Table~\ref{tab:claimcarrying}. \textit{Lexical+DNLI} performs the best and all baselines have low F1 scores for \textsc{Non-Verifiable} claims.

\subsection{Evidence Retrieval}
Evidence retrieval consists of two steps: 1) Document Retrieval, 2) Evidence Sentence selection.

\subsubsection{Document Retrieval}
We test two methods for document retrieval: The first one is \textit{WikiAPI}\footnote{\url{www.github.com/UKPLab/
fever-2018-team-athene}}, which retrieves Wikipedia pages and is used in past fact-checking work~\cite{hanselowski-etal-2018-ukp, stammbach-neumann-2019-team, liu-etal-2020-fine}. It uses the AllenNLP constituency parser~\cite{gardner-etal-2018-allennlp} to extract potential entities from the claims. Then it feeds the entities as queries through the MediaWiki API$^{\ref{note1}}$ and returns up to three Wikipedia pages per query. For each Wikipedia page, we query the KILT~\cite{petroni-etal-2021-kilt} knowledge source to get the first 5 paragraphs of the page. 
We create two versions of this method: a) \textit{Wiki-ctx} which concatenates the last two turns of the dialogue context with the response claim before document retrieval and b) \textit{Wiki-claimonly} - which uses just the claim. 
The second method is \textit{Dense Passage Retrieval (DPR)}~\cite{karpukhin-etal-2020-dense}, a dual encoder based model which retrieves documents using BERT~\cite{devlin-etal-2019-bert} trained by metric learning. We create three versions of this method: a) \textit{DPR-original}, which uses the original DPR trained on question-answering tasks, b) \textit{DPR-WoWft-claimonly}, which is fine-tuned on the WoW dataset to retrieve documents relevant to a query composed only of a response claim, and c) \textit{DPR-WoWft-ctx}, which is also fine-tuned on WoW dataset but uses both the context as well as the response as a query (training details are provided in Appendix~\ref{sec:implementation}). For DPR-based methods we retrieve the top 100 documents. A document is relevant if it contains a gold evidence sentence.

\begin{table}[t]
\centering
\small
\begin{tabular}{lcc}
\toprule
Model& Recall\\ \hline
DPR-original&40.3\\
DPR-WoWft-claimonly&44.7\\
DPR-WoWft-ctx&58.8\\
\hline
Wiki-claimonly&60.8\\
Wiki-ctx&\textbf{75.0}\\
\bottomrule
\end{tabular}
    \caption{Document recall for the test set. Incorporating dialogue context in document improves performance on both WikiAPI and DPR.}
    \vspace{-0.5pc}
    \label{tab:documentretrieval}
\end{table}

We present the document recall results in Table~\ref{tab:documentretrieval}. WikiAPI methods outperform DPR-based methods.
Both methods show better performance when dialogue context is used in retrieval. DPR is typically able to retrieve documents with the correct topic but often fails to retrieve a relevant evidence sentence.
Entity linking is crucial for fact-checking in dialogue and WikiAPI is able to leverage that capability for better performance. 

\begin{table}[t]
\centering
\small
\begin{tabular}{lcc}
\toprule
                  & \multicolumn{2}{c}{Recall@5} \\
\hline
Model   & DPR-WoWft-ctx  & Wiki-ctx  \\
\hline

Ret-only-claim           & 67.1           & 70.1      \\
Ret-with-context         & \textbf{69.3}           & \textbf{75.4}     \\
\bottomrule
\end{tabular}
\caption{Evidence sentence Recall@5 for the test set.}
    \vspace{-0.5pc}
    \label{tab:evidenceretrieval}
\end{table}

\begin{table*}[t]
\centering
\small
\begin{tabular}{lcccccc}
\toprule
& \multicolumn{2}{c}{Oracle-Evidence} & \multicolumn{2}{c}{Wiki-Evidence} & \multicolumn{2}{c}{DPR-Evidence} \\ \hline
Model               & Accuracy         & Macro F1         & Accuracy          & Macro F1          & Accuracy        & Macro F1       \\ \hline
 \noalign{\vskip 1pt}
DNLI                 & 43.3                         & 35.4                         & 39.1                         & 31.5                         & 38.4                         & 29.5                         \\
DECODE               & 37.8                         & 30.3                         & 35.3                         & 25.3                         & 34.5                         & 22.5                         \\
VitaminC             & 57.6                         & 56.1                         & 46.2                         & 44.7                         & 45.9                         & 44.2                         \\
CorefBert-Colloquial & 61.4                         & 60.0                         & 47.6                         & 45.2                         & 46.4                         & 41.1                         \\
Colloquial           & 63.5                         & 62.8                         & 48.1                         & 46.3                         & 48.7                         & 46.4                         \\
Aug-WoW & \textbf{69.2} & \textbf{69.0} & \textbf{51.6} & \textbf{51.3} & \textbf{51.5} & \textbf{50.2} \\ 
\bottomrule
\end{tabular}
    \caption{Results for claim verification on the test set. We experiment with three types of evidences and report Accuracy and Macro F1 scores in percentage. \modelname outperforms all baselines across all settings.}
    \label{tab:verification}
\end{table*}

\begin{table*}[t]
\centering
\small
\begin{tabular}{lcccccc}
\toprule
& \multicolumn{2}{c}{Oracle-Evidence} & \multicolumn{2}{c}{Wiki-Evidence} & \multicolumn{2}{c}{DPR-Evidence} \\ \hline
Model               & Accuracy         & Macro F1         & Accuracy          & Macro F1          & Accuracy        & Macro F1       \\ \hline
 \noalign{\vskip 1pt}    
Aug-WoW-noctx              & {68.1}                         & {68.1}                         & \textbf{52.4}                         & \textbf{52.3}                         & \textbf{52.4}                         & \textbf{51.3}                        \\
Aug-WoW-BertLarge & \textbf{70.9} & \textbf{70.9} & 45.8 & 44.6 & 43.5 & 39.1 \\
Aug-WoW & {69.2} & {69.0} & {51.6} & {51.3} & {51.5} & {50.2} 
\\\bottomrule
\end{tabular}
    \caption{Results for claim verification on the test set with Aug-WoW model ablations.}
    \vspace{-0.5pc}
    \label{tab:ablationaugwow}
\end{table*}

\subsubsection{Evidence Sentence Selection}
In evidence sentence selection, a final set of top $k$ evidence sentences are chosen from the set of documents $D_c$ retrieved in the previous step for claim $c$. First, we create a candidate evidence sentence set $S_c$ by taking the union of all sentences in $D_c$.
We fine-tune a Bert-base model for ranking the candidate sentences in $S_c$. The model is trained to predict -1 for irrelevant evidence and 1 for relevant evidence for a given claim. 
We use the context-response pairs from the WoW dataset for training the model. Besides using randomly selected evidence sentences, to create hard negative examples for training, we also chose sentences from the set of articles $K_i$ shown to the wizard during WoW data collection. These sentences are close in content and topic to the gold evidence sentence and form hard negative candidates for the model. At test time, we use the evidence sentences in the top $k$ rank with a score of more than 0. Similar to document retrieval, we created two versions of the model: 1) Ret-with-context, and 2) Ret-only-claim, based on whether the last two utterances of the dialogue context were included in the input to the BERT model. We present the performance of the models in Table~\ref{tab:evidenceretrieval} for two of the best performing document retrieval models Wiki-ctx and DPR-WoWft-ctx. 
We find that recall@5 values for both models are higher when dialogue context is added as an input with the claim.

\subsection{Claim Verification}
\label{sec:claimverification}
In claim verification, a claim $c$ is classified as \textsc{Supported}, \textsc{Refuted}, or \textsc{NEI} given a context $C$ and evidence sentences set $S_c$. 

\subsubsection{Baselines}
\label{sec:baselines}
\noindent
\textbf{DNLI}~\cite{welleck-etal-2019-dialogue} Dialogue NLI dataset  contains sentence pairs labeled as entailment, neutral, or contradiction derived from dialogues. Entailment maps to \textsc{Supported}, neutral maps to \textsc{NEI}, and contradiction maps to \textsc{Refuted} in our task. We train a Bert-base model on their training set of 310,110 data points. 

\noindent
\textbf{DECODE}~\cite{nie-etal-2021-like} Dialogue Contradiction Detection dataset contains both human-human and human-bot contradictory dialogues. The train set contains 27,948 data points with two labels contradiction and non-contradiction. We train a Bert-base model with the last two utterances of the context and the response as input to the model.

\noindent
\textbf{VitaminC}~\cite{schuster-etal-2021-get} VitaminC is a large-scale fact verification dataset which is based on contrastive claim-evidence pairs created from Wikipedia edits. They train models that avoid claim-only biases and are more sensitive to changes in the evidence. 
We use their ALBERT-base model finetuned on FEVER~\cite{thorne-etal-2018-fever} and their VitaminC dataset.

\noindent
\textbf{Colloquial}~\cite{kim-etal-2021-robust} It contains colloquial claims converted from FEVER dataset claims into colloquial style. 
It has 410k colloquial claim-evidence pairs in the training set and is well aligned to our task because of its colloquial nature. We fine-tune a Bert-base model on this dataset.

\noindent
\textbf{CorefBert-Colloquial}~\cite{ye-etal-2020-coreferential} is one of the best performing models on FEVER and is designed to better capture and represent the coreference information. We use their model which uses kernel graph attention network (KGAT)~\cite{liu-etal-2020-fine} and fine-tune it on Colloquial claims.

\noindent
\textbf{\modelname} We propose a novel model which is trained on weakly supervised training data. \dataname is meant to be used only for validation and test, and we do not train a model on \dataname to avoid creating a model which can simply learn to solve the dataset instead of the task.
Instead, we leverage the techniques described in section~\ref{sec:augmentation} to create synthetic training data for each category of claims. 
For \textsc{Supported} claims, we use the claim-evidence pair from the original WoW dataset. We use the \textit{Lexical} baseline from section~\ref{sec:claimdetection} to filter out Non-Verifiable claims, which leads to 46,934 \textsc{Supported} claims. 
We follow the methods \textit{Negation} and \textit{Substitution} from section~\ref{sec:augmentation} to create 38,895 \textsc{Refuted} claims.
We create \textsc{NEI} claims using two methods: 1) For every context-claim-evidence triplet, we substitute the evidence with random unrelated evidence. 2) We use the \textit{Generation} approach from section~\ref{sec:augmentation} to condition the generation on random evidence. We select a subset of 40,000 NEI claims from the two approaches. We fine-tune the \textit{Colloquial} baseline model on this synthetic dataset. The input to the model is the sequence of the last 2 context utterances separated by [EOT] token, followed by the claim.

For all Bert-based models, all evidence sentences are concatenated together. More details about training the baselines are provided in Appendix~\ref{sec:implementation}.

\begin{table}[t]
\centering
\addtolength{\tabcolsep}{-2.5pt}
\small
\begin{tabular}{lcccc}
\toprule
                     & \multicolumn{2}{c}{Generated} & \multicolumn{2}{c}{Written} \\ \hline
Model                & Accuracy         & Macro F1         & Accuracy          & Macro F1          \\ \hline
 \noalign{\vskip 1pt}
DNLI                 & 50.9             & 38.4             & 34.8              & 31.0              \\
DECODE               & 36.5             & 30.4             & 39.3              & 30.1              \\
VitaminC             & 48.9             & 42.1             & 60.8              & 60.3              \\
\begin{tabular}[c]{@{}l@{}}CorefBert-\\ Colloquial\end{tabular} & 56.9             & 51.6             & 66.4              & 65.5              \\
Colloquial           & 61.3             & 56.9             & 64.7              & 64.6              \\
Aug-WoW              & \textbf{63.9}             & \textbf{60.7}             & \textbf{74.2}              & \textbf{74.0}   \\
\bottomrule
\end{tabular}
    \caption{Results for claim verification on the test set for Generated and Written claims. }
    \vspace{-1.0pc}
    \label{tab:generatedwritten}
\end{table}

\begin{table*}[t]
\small
\centering
\begin{tabular}{lp{8.4cm}p{4.8cm}}
\toprule
Context  & Biathlon means two sports right? What is the other sport?                                             & \multirow{3}{*}{\shortstack[l]{Response type: Generated\\DNLI: S, CorefBERT-Colloquial: S, \\DECODE: R,  VitaminC: NEI,\\Colloquial: S, AugWoW: R,\\ Human: R}}\Bstrut \\
Response & Biathlon combine the two sports into one event called the cross country ski race. It's a lot of fun! &                                                                                                     \\
Evidence & Biathlon: The biathlon is a winter sport that combines cross-country skiing and rifle shooting.       &                                                                                                     \\

\hline
\Tstrut
Context  & Do you know if professional cheerleaders make a lot of money?                                            &  \multirow{3}{*}{\shortstack[l]{Response type: Generated\\DNLI: S, CorefBERT-Colloquial: NEI, \\DECODE: R, VitaminC: S,\\Colloquial: S, AugWoW: NEI,\\ Human: NEI}} \\
Response & The whole point of cheerleading is to show off their skills, so I'm sure they get paid a lot of money. &       \\ 
Evidence & Cheerleading: Cheerleading originated in the United States with an estimated 1.5 million participants in all-star cheerleading.       &                                                                                                     \\
\hline
\Tstrut
Context  & Japanese is even harder, the language is difficult to speak.                                            &  \multirow{3}{*}{\shortstack[l]{Response type: Generated\\DNLI: S, CorefBERT-Colloquial: NEI, \\DECODE: S, VitaminC: NEI,\\Colloquial: NEI, AugWoW: NEI,\\ Human: R}}\Bstrut \\
Response & The origins of the language lie in the prehistoric times when many cultures spoke to one another. &       \\ 
Evidence & Japanese language: Little is known of the language's prehistory, or when it first appeared in Japan.       &                                                                                                     \\
\hline
\Tstrut
Context  & I might recognize if I heard it.  Who else did you listen to in the 90s?                                          & 
\multirow{3}{*}{\shortstack[l]{Response type: Written\\DNLI: S, CorefBERT-Colloquial: R, \\DECODE: R, VitaminC: NEI,\\Colloquial: R, AugWoW: R,\\ Human: S}} \\
Response & I also listened to another group Dave Grohl was apart of called Them Crooked Vultures. It was not one of his best groups. &       \\ 
Evidence & Dave Grohl: He is the drummer and co-founder of the rock supergroup Them Crooked Vultures.     &                                                                                                     \\
\bottomrule
\end{tabular}
    \caption{Sample dialogue contexts, claims, evidences and model predictions. We also indicate whether the response is automatically generated or human written. Here S stands for \textsc{Supported} and R for \textsc{Refuted}. }
    \label{tab:samples}
    \vspace{-5mm}
\end{table*}

\subsubsection{Results}
\label{sec:resultsverification}
Table \ref{tab:verification} summarizes the results for claim verification on the test set. \textsc{Non-Verifiable} claims are included in the NEI category.
We experiment with three evidence retrieval settings - 1) Oracle Evidence, where we use gold evidence, 2) Wiki-Evidence, where we use Wiki-ctx for document retrieval and Ret-with-context for evidence selection, and 3) DPR-Evidence, where we use DPR-WoWft-ctx for document retrieval and Ret-with-context for evidence selection. We set the maximum evidence to 5. In all three settings, Aug-WoW outperforms baselines and the performance of all baselines drops when retrieved evidence is used compared to when oracle evidence is used. This indicates that evidence retrieval is an important step for this task. Even with oracle evidence, none of the models achieve an accuracy higher than 70\%, which leaves abundant opportunity for future improvements.
Colloquial baseline is the closest to Aug-WoW since it has been trained on conversation-like colloquial claims.
Although Colloquial and CorefBert-Colloquial perform better than VitaminC with oracle evidence, the contrastive nature of VitaminC helps it perform better with retrieved evidences. 

In Table~\ref{tab:generatedwritten}, we present the claim verification results on the Test set using oracle evidence on Generated and Written claims separately. The performance of all models is lower on Generated claims compared to Written claims. This is expected since as we mentioned in ``Final claim set creation'' in section~\ref{sec:augmentation}, the Generated claims were chosen from a larger candidate claims set based on the difficulty of existing models to classify those claims. Thus Generated claims in \dataname are more challenging. Furthermore, Aug-WoW's performance is high on both types of claims, however, the gain in its performance on Written claims is higher on Written claims compared to Generated claims.

In Table~\ref{tab:ablationaugwow}, we present the claim verification results on the test set with Aug-WoW model ablations. In Aug-WoW-noctx we do not concatenate the dialogue context, and in Aug-WoW-BertLarge we use the Bert-Large model as base architecture.
Aug-WoW-noctx is comparable to Aug-WoW, and has slightly lower performance with Oracle evidence.
Although Aug-WoW-BertLarge performs better with oracle evidence, it is more sensitive to the evidence quality and performs poorly with retrieved evidence. 

To test if a model that relies solely on claims and no evidence can leverage lexical biases in the claims to obtain good performance on \dataname, we train a model \textit{Aug-WoW-claimonly} with no evidence included during training and testing. \textit{Aug-WoW-claimonly} achieves 33.2\% accuracy and 28.9\% macro F1 score on the \dataname test set. Thus, a model can not exploit lexical cues in the claims of \dataname to obtain good performance. 

We report performance on a two-way classification experiment in Appendix~\ref{sec:otherresults} (Table~\ref{tab:verification2way}) where we combine \textsc{Refuted} and \textsc{NEI} into a single class named \textsc{Not-Supported}.

\subsubsection{Discussion}
We present sample dialogue contexts, claims, oracle evidence for the claims along with model predictions in Table~\ref{tab:samples}. We found that models tend to incorrectly predict a \textsc{Refuted} or \textsc{NEI} response as \textsc{Supported} when there is significant overlap between the evidence and the claim while ignoring the semantics. The first example illustrates this point where the presence of terms ``biathlon'' and ``cross country skiing'' misleads some models to predict \textsc{Supported} incorrectly. Similarly, models predict \textsc{Supported} or \textsc{Refuted} for a \textsc{NEI} claim due to word overlap between claim and evidence, as shown in the second example. Models also often fail to perform complex and commonsense-based reasoning during verification. In the third example, although humans can reason that the claim is \textsc{Refuted} by the evidence, all models fail to correctly classify the claim. Finally, models struggle with lexical biases and separating the colloquial part of a claim from its factual parts. In the fourth example, although there is significant overlap between the claim and the evidence, models are fooled by the presence of the word ``not one of'', and predict a \textsc{Supported} claim as \textsc{Refuted}.

\section{Conclusion}

We propose a new benchmark, {\dataname}, for fact-checking in dialogue created based on grounded dialogues from the Wizard-of-Wikipedia dataset. Besides human-written response claims, we also create synthetic claims with operations such as contradiction, infilling and substitutions. We hire qualified crowd workers to annotate responses into \textsc{Non-Verifiable, supported, refuted}, or \textsc{NotEnoughInformation} categories along with corresponding evidence. We point out empirically that existing fact-checking models trained on non-dialogue data fail to perform well on our task. We demonstrate how to leverage automatically generated responses as weak supervised signals to improve performance. We hope that {\dataname} can facilitate fact-checking, and consistency modeling and evaluation research in the dialogue community.

\section*{Ethical Considerations \& Broader Impact}
In this paper, we study the problem of fact-checking in dialogue. The \dataname benchmark dataset proposed in this work could be helpful in creation of more accurate automatic fact checking systems and metrics, and ultimately creation of dialogue systems which are more faithful to factual knowledge and are thus more trustworthy. 
Automatic fact-checking of dialogue could be useful in many real-life scenarios where  conversations need to be properly monitored to avoid spread of misinformation and disinformation, and where the conversation participants are needed to be given accurate information. 
However, \dataname benchmark only covers a specific domain with Wikipedia as background knowledge. Furthermore, even with our best efforts to ensure high quality and accuracy, the dataset might still contain incorrect labels and biases in some instances.
This could pose a risk if models that are evaluated or built using this benchmark are used in domains not covered by the dataset or if they leverage evidence from unreliable or biased resources.
Thus the proposed benchmark should not be treated as a universal tool for all domains and scenarios.
In our work, we mitigate this risk by using the trusted source of Wikipedia for evidence and by curating hard training and testing instances using automated generation approaches.
Considerable additional work is needed to improve the scope, coverage and validity of fact-checking systems and metrics, but our work provides a cautious yet concrete step towards developing fact checking systems for dialogue. 
training and testing instances using automated generation approaches.

\bibliography{anthology,custom}
\bibliographystyle{acl_natbib}

\appendix

\begin{table*}[t]
\centering
\small
\begin{tabular}{lcccccc}
\toprule
& \multicolumn{2}{c}{Oracle-Evidence} & \multicolumn{2}{c}{Wiki-Evidence} & \multicolumn{2}{c}{DPR-Evidence} \\ \hline
Model                & Accuracy         & Macro F1         & Accuracy          & Macro F1          & Accuracy        & Macro F1       \\
DNLI                 & 42.0                         & 34.9                         & 39.0                         & 31.1                         & 38.2                         & 30.1                         \\
DECODE               & 31.6                         & 29.2                         & 33.5                         & 25.7                         & 31.1                         & 21.2                         \\
VitaminC             & 60.5                         & 58.4                         & 45.2                         & 43.8                         & 46.1                         & 44.2                         \\
CorefBert-Colloquial & 64.5                         & 63.0                         & 46.8                         & 44.4                         & 46.2                         & 42.4                         \\
Colloquial           & 65.0                         & 63.1                         & 48.6                         & 46.5                         & 51.3                         & 48.4                         \\
Aug-WoW              & \textbf{70.4}                         & \textbf{70.4}                         & \textbf{51.2}                         & \textbf{51.1}                         & \textbf{50.4}                         & \textbf{49.6}    \\
\bottomrule
\end{tabular}
    \caption{Results for claim verification on the validation set. We experiment with three types of evidences and report Accuracy and Macro F1 scores in percentage. \modelname outperforms all baselines across all settings.}
    \vspace{-0.5pc}
    \label{tab:verificationvalid}
\end{table*}


\begin{table*}[t]
\centering
\small
\begin{tabular}{lcccccc}
\toprule
& \multicolumn{2}{c}{Oracle-Evidence} & \multicolumn{2}{c}{Wiki-Evidence} & \multicolumn{2}{c}{DPR-Evidence} \\ \hline
Model                & Accuracy         & Macro F1         & Accuracy          & Macro F1          & Accuracy        & Macro F1       \\
DNLI                 & 43.8                         & 33.7                         & 41.3                         & 32.2                         & 41.3                         & 30.4                         \\
DECODE               & 41.8                         & 31.7                         & 39.0                         & 26.7                         & 38.1                         & 23.8                         \\
VitaminC             & 52.7                         & 52.9                         & 41.3                         & 40.8                         & 41.1                         & 40.9                         \\
CorefBert-Colloquial & 64.1                         & 61.9                         & 50.1                         & 46.5                         & 50.0                         & 43.0                         \\
Colloquial           & 63.4                         & 62.3                         & 48.1                         & 45.9                         & 49.8                         & 46.3                         \\
Aug-WoW              & \textbf{69.7}                         & \textbf{69.0}                         & \textbf{51.7}                         & \textbf{50.5}                         & \textbf{52.8}                         & \textbf{49.6}        \\
\bottomrule
\end{tabular}
    \caption{Results for claim verification on the test set for 3-way classification where Non-Verifiable claims with NEI-Personal labels are removed and for NEI only Verifiable claims are kept. We report Accuracy and Macro F1 scores in percentage. }
    \vspace{-0.5pc}
    \label{tab:verification3wayNEIfactual}
\end{table*}

\begin{table*}[t]
\centering
\small
\begin{tabular}{lcccccc}
\toprule
& \multicolumn{2}{c}{Oracle-Evidence} & \multicolumn{2}{c}{Wiki-Evidence} & \multicolumn{2}{c}{DPR-Evidence} \\ \hline
Model                & Accuracy         & Macro F1         & Accuracy          & Macro F1          & Accuracy        & Macro F1       \\
DNLI                 & 48.1                         & 46.5                         & 47.2                         & 46.3                         & 43.9                         & 42.0                         \\
DECODE               & 65.4                         & 62.5                         & 63.2                         & 52.2                         & 62.3                         & 47.1                         \\
VitaminC             & 76.2                         & 67.7                         & 70.6                         & 60.8                         & 69.8                         & 61.6                         \\
CorefBert-Colloquial & 72.3                         & 71.8                         & 63.3                         & 62.9                         & 57.7                         & 57.7                         \\
Colloquial           & 76.8                         & 75.2                         & 66.4                         & 65.1                         & 63.5                         & 63.0                         \\
Aug-WoW              & \textbf{80.6}                         & \textbf{78.8}                         & \textbf{69.0}                         & \textbf{67.4}                         & \textbf{68.2}                         & \textbf{67.3}      \\
\bottomrule
\end{tabular}
    \caption{Results for claim verification on the test set for 2-way classification - \textsc{Supported} and \textsc{Not-Supported}. We combine \textsc{Refuted} and \textsc{NEI} into \textsc{Not-Supported}. We report Accuracy and Macro F1 scores in percentage. }
    \vspace{-0.5pc}
    \label{tab:verification2way}
\end{table*}


\section{Supplementary Results}
\label{sec:otherresults}
We present the claim verification results on the validation set in Table~\ref{tab:verificationvalid}. The trend in performance is similar to the trend observed in the test set reported in \ref{tab:verification}.
In our human studies discussed in subsection \textit{Data validation} of section~\ref{sec:quality}, we observe that workers confuse between \textsc{Refuted} and \textsc{NEI} labels. Furthermore, there are cases where the workers can miss finding an evidence which refutes a claim on Wikipedia and label the claim as NEI even though they are instructed to find and verify a claim by visiting Wikipedia. Similar findings were reported in other fact-checking tasks~\cite{jiang-etal-2020-hover}. Hence we perform another experiment where we combine \textsc{Refuted} and \textsc{NEI} into a single class, and name it \textsc{Not-Supported}. We present the claim verification results on test set for this setting in Table~\ref{tab:verification2way}. The performance of all baselines is higher since the task is transformed to a 2-way classification task from a 3-way classification task. Aug-WoW performs the best in this setting.

In Section\ref{sec:resultsverification}, we discuss results where \textsc{Non-Verifiable} claims are included in the NEI category. In Table~\ref{tab:verification3wayNEIfactual}, we present the results for 3-way classification on test set where \textsc{Non-Verifiable} claims with \textsc{NEI-Personal} labels are removed, that is, only Verifiable claims are kept for NEI labelled claims. The trends in results are similar to the ones observed in Table~\ref{tab:verification}.



\begin{figure}[t]
    \centering
    \includegraphics[width=.6\linewidth]{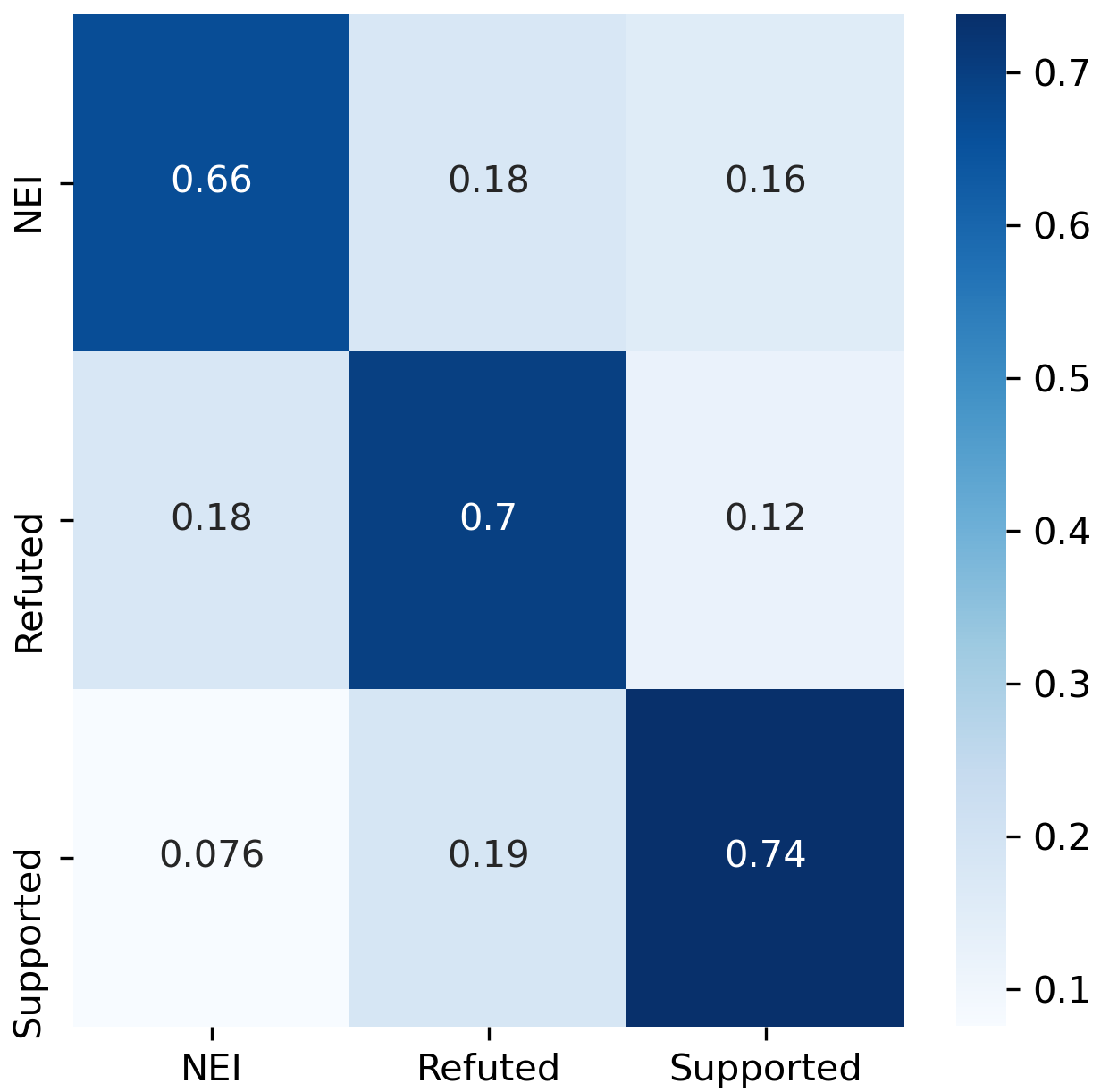}
    \vspace{-.5pc}
    \caption{The Confusion matrix of Aug-WoW model.}
    \label{fig:confusion}
\end{figure}

We show the confusion matrix of our Aug-WoW model in Figure~\ref{fig:confusion}. 
Aug-WoW has the lowest performance on NEI claims and highest confusion between NEI and Refuted classes.

\begin{table*}[t]
\centering
\small
\begin{tabular}{l|l} 
\toprule
Context & \begin{tabular}[c]{@{}l@{}}A: I prefer to eat fish that is not farm raised due to the pesticides in the food.\\B: Yes the two most common are atlanticcod and pacific cod\\A:~Most cod sold in stores is farm raised, and also the cod you eat in restaurants.\end{tabular} \\ 
\hline
\begin{tabular}[c]{@{}l@{}}Responses\\(Generated)\end{tabular} & \begin{tabular}[c]{@{}l@{}}Response 1:~There are other varieties of cod as well, like the black, red, white, and yellow\\Evidence:~Cod flesh is moist and flaky when cooked and is white in colour. It change colour at certain \\water depths. It has two distinct colour phases: gray-green and reddish brown\\Labels: Factual, Refuted\\\\Response 2:~I read that it is a popular food with a mild flavor and a dense flaky flesh\\Evidence:~Cod is popular as a food with a mild flavour and a dense, flaky white flesh.\\Labels: Factual, Supported\\~~\\Response 3:~I read that it is a large fruit with a bunch flower and a fleshy petals.\\Evidence:~Cod is popular as a food with a mild flavour and a dense, flaky, white flesh.\\Labels: Factual, Refuted\end{tabular} \\ 
\toprule
Context & \begin{tabular}[c]{@{}l@{}}A: Elvis's first RCA single was "Heartbreak Hotel" released in 1956 and became a number one hit in US.\\B: Right, he became popular pretty quickly! When did he die?\end{tabular} \\ 
\hline
\begin{tabular}[c]{@{}l@{}}Responses\\(Written)\end{tabular} & \begin{tabular}[c]{@{}l@{}}Response 1: Some think he died August 16, 1977. He helped pioneer the popular sound of rock and roll.\\Evidence:~~Elvis Aaron Presley (January 8, 1935~– August 16, 1977) was an American singer, musician,\\and actor.~He became the leading figure of the newly popular sound of rock and roll.\\Labels: Factual, Supported~~\\\\Response 2: Some think he died August 25, 1988. He helped pioneer the popular sound of rap music.\\Evidence:~~Elvis Aaron Presley (January 8, 1935~– August 16, 1977) was an American singer, musician,\\and actor.~He became the leading figure~of the newly popular sound of rock and roll.\\Labels: Factual, Refuted~~\\\\Response 3: I am trying to remember when he died. But most people in Russia see him as an idol.\\Evidence:~Elvis Presley - He became the leading figure~of the newly popular sound of rock and roll.\\Labels: Factual, NEI~~\end{tabular} \\
\bottomrule
\end{tabular}
\caption{We present two examples from DialFact dataset: The top context has responses which were automatically generated and then labelled. The bottom context has responses written and then labelled. The labels and evidence are shown below the responses.}
\label{tab:examples}
\vspace{-2mm}
\end{table*}

\section{Implementation Details}
\label{sec:implementation}
First we discuss the implementation details for claim generation techniques in section~\ref{sec:augmentation}. For Negation we use the implementation from fever-2 baseline\footnote{\url{www.github.com/j6mes/fever2-baseline}}~\cite{thorne-etal-2019-evaluating}. For the T5 model in \textit{Mask-and-Fill} and Blenderbot model in \textit{Generation} approach, we use the models and training scripts available in the Hugging Face's Transformers repository\footnote{\url{www.github.com/huggingface/transformers/}}. Blenderbot was finetuned on full WoW training dataset with batch size of 40.

We next discuss the implementation details for the document retrieval methods. For WikiAPI method, \citet{kim-etal-2021-robust} pointed out that WikiAPI method naively retrieves documents related to filler words such as ``I'', ``Yes'', ``They'' etc. frequently. In our implementation of WikiAPI we mitigate this issue by filtering out such colloquial phrases by using a manually created stopwords list. We remove the stopwords from the candidate set of entities on which MediaWiki API is called. Our experiments showed significant improvement in the quality of the returned documents. For DPR, we use the \textit{wiki\_dpr} dataset available in the Hugging Face Datasets library\footnote{\url{www.huggingface.co/datasets/wiki_dpr}} for document retrieval. It contains 21M passages from wikipedia along with their DPR embeddings. The wikipedia articles are split into multiple, disjoint text blocks of 100 words as passages. We retrieve top 100 documents per claim. We finetune the claim encoders for \textit{DPR-WoWft-claimonly} and \textit{DPR-WoWft-ctx} using the original DPR implementation\footnote{\url{www.github.com/facebookresearch/
DPR}}. The original biencoder was trained on natural questions dataset. We only fine-tune the question encoder of the DPR model. DPR training data consists of positive, random negatives and hard negative pairs. For positive claim-evidence document pairs, we use the response-knowledge sentence pairs in the original WoW dataset, where we filter out \textsc{Non-Verifiable} claims using the \textit{Lexical} baseline from section~\ref{sec:claimdetection}. For hard negatives, we follow the instructions in the DPR repository and mine hard negatives using the original DPR index and encoder (facebook/dpr-question\_encoder-single-nq-base) itself. Specifically, we use DPR to retrieve top 2 evidences per claim and use them as a hard negative if they are not the same as the original knowledge sentence for the claim in the WoW dataset. We finetune the base DPR encoder on the aforementioned constructed data and convert only the question encoder checkpoints into Hugging Face model format. Since the Wikipedia version used for evidence in WoW dataset (and hence in DialFact evidence), and Hugging Face's wiki\_dpr (used for document retrieval in our experiments) are different, even if WikiAPI and DPR methods retrieve a correct document, it might not exactly match the evidence we picked up from WoW dataset due to wording changes and edits between the two versions of Wikipedia pages. Therefore we relax the requirements from exact document matching to partial matching. That is, we assume a retrieved document matches a gold document if either the initial half or final half of the retrieved document matches the gold evidence document's half. 

We next discuss the implementation details for the models for claim verification~\ref{sec:claimverification}. For VitaminC, we use the tals/albert-base-vitaminc-fever model available in their repo\footnote{\url{www.github.com/TalSchuster/VitaminC}}. We finetune
CorefBERT-base for CorefBERT and use the official code from the authors\footnote{\url{www.github.com/thunlp/CorefBERT/tree/master/FEVER}}. We train AugWoW and Colloquial models using the code from the VitaminC repo\footnote{\url{www.github.com/TalSchuster/VitaminC}} on a machine with 4 NVIDIA A100 GPUs and train batch size of 100. We use the validation set performance for model selection.

\begin{figure*}[tb]
    \centering
    \vspace{-.2pc}
    \includegraphics[width=0.98\textwidth]{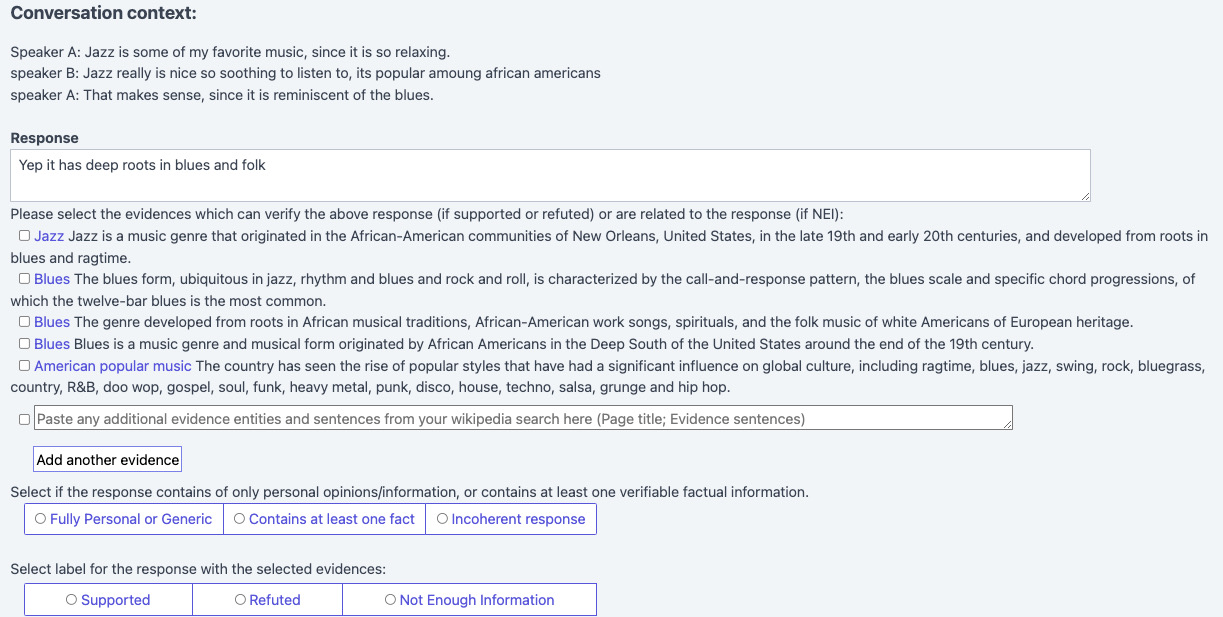}
    \vspace{-.5pc}
    \caption{Annotation interface for claim labeling. Workers are shown a conversation context, a claim or response to the context, and evidence sentences from Wikipedia related to the response. They are asked to add any additional evidence necessary for labelling. They first select if the response is \textsc{Verifiable} or \textsc{Non-Verifiable}. Then they select one of the categories - \textsc{Supported, Refuted and Not Enough Information}.  }
    \label{fig:interface1}
\end{figure*}

\section{AMT Instructions}
\label{sec:amt_info}
We present the screenshot of the annotation interface is shown in Figure~\ref{fig:interface1}. Workers were paid an avergae of \$8-10 per hour across all tasks.
For the claim labelling task, workers were told that they will be shown a conversation between two speakers, some previously created responses to the conversation, and some Wikipedia knowledge snippets related to the response (which we will call evidence henceforth). They will label some dialogue responses which could belong to one of the 3 categories mentioned below.

\noindent \textbf{Supported}: The response should exclusively use factual information which can be verified by the given evidence sentences and is correct or true in light of the evidence. A response is verifiable if evidence could be retrieved from Wikipedia, which decreases the uncertainty about the truthfulness (or falsehood) of the statement.

\noindent Example 1:
\begin{itemize}[leftmargin=*]
    \item Context: I think Jazz is an American creation!
    \item Evidence: Jazz has roots in West African cultural and musical expression, and in African-American music traditions including blues and ragtime, as well as European military band music.
    \item Response: Its roots include African-American music traditions including blues and ragtime
    \item Explanation: Response is natural and can be verified from the evidence.
\end{itemize}

\noindent Example 2:
\begin{itemize}[leftmargin=*]
    \item Context: What are the three different waterfalls Niagra is made from? Can you please share with me?
    \item Evidence: From largest to smallest, the three waterfalls are the Horseshoe Falls, the American Falls, and the Bridal Veil Falls.
    \item Response: The three waterfalls are the Horseshoe Falls, the American Falls and the Bridal Veil Falls.
    \item Explanation: Response is natural and can be verified from the evidence as all facts mentioned are correct.
\end{itemize}

\noindent \textbf{Refuted}: The response contains factual information which is ``incorrect'' or ``false'' in light of the evidence, that is it contradicts the evidence. The response should be marked refuted if even a small part of the response is incorrect.

\noindent Example 1:
\begin{itemize}[leftmargin=*]
    \item Context: I think Jazz is an American creation!
    \item Evidence: Jazz has roots in West African cultural and musical expression, and in African-American music traditions including blues and ragtime, as well as European military band music.
    \item Response: Its roots include American music traditions including blues and ragtime
    \item Explanation: Roots are African-American, not American.
\end{itemize}

\noindent Example 2:
\begin{itemize}[leftmargin=*]
    \item Context: What are the three different waterfalls Niagra is made from? Can you please share with me?
    \item Evidence: From largest to smallest, the three waterfalls are the Horseshoe Falls, the American Falls and the Bridal Veil Falls.
    \item Response: The three waterfalls are the Horseshoe Falls, the American Falls and the Sommer Falls.
    \item Explanation: One of the falls is incorrect based on the evidence.
\end{itemize}

\noindent \textbf{Not Enough Information}: The response can not be verified (supported or refuted) with Wikipedia evidence. Moreover, for this response, it is allowed to use information/knowledge that might not be available in Wikipedia but you assume to be general knowledge, e.g. that 90s refers to the time span from 1990 to 1999.

\noindent Example 1:
\begin{itemize}[leftmargin=*]
    \item Context:  I think Jazz is an American creation!
    \item Evidence: Jazz has roots in West African cultural and musical expression, and in African-American music traditions including blues and ragtime, as well as European military band music.
    \item Response: Jazz is now played in all parts of the world except Russia.
    \item Explanation: The response is not a personal opinion and the provided evidence can’t be used to verify the stated fact.
\end{itemize}

\noindent Example 2:
\begin{itemize}[leftmargin=*]
    \item Context: What are the three different waterfalls Niagra is made from? Can you please share with me?
    \item Evidence: From largest to smallest, the three waterfalls are the Horseshoe Falls, the American Falls and the Bridal Veil Falls.
    \item Response: I think three waterfalls all intersect multiple times. I am trying to remember the names.
    \item Explanation: The stated fact can not be verified from the evidence.
\end{itemize}

\noindent We ask workers to do the following:
\begin{itemize}[leftmargin=*]
    \item Read the context carefully and if writing or editing a response, write minimum of 9 words.
    \item The label should be exclusively based on the response and the selected evidence sentences.
\end{itemize}

\noindent We ask workers to NOT do the following:
\begin{itemize}[leftmargin=*]
    \item While writing or editing a response please avoid typos and mis-spelling as much as possible.
    \item While writing or editing a response, do not use “know-it-all” phrases such as "did you know" in your responses - e.g., the response "did you know that the Berlin Wall was demolished in 1989" will not be accepted.
\end{itemize}

\noindent \textbf{Personal/generic response}: We give workers some examples of personal response. The response should not make any factual claim that could be verified using Wikipedia or any knowledge source. It can contain facts that are personal opinions or background of the speaker, but no fact pertinent to encyclopedic knowledge. The response should be a good follow-up to the conversation.

\noindent Example 1:
\begin{itemize}[leftmargin=*]
    \item Context: I do not understand why some people enjoy hunting.
    \item Evidence: Hunting is the practice of killing or trapping animals.
    \item Response 1: I enjoy going out in the woods to hunt animals.
    \item Response 2: Wow interesting. I have mostly used hunting as a means of pest control.
    \item Explanation: Even if hunting can be used as pest control, it is a personal detail or opinion here.
\end{itemize}

\noindent Example 2:
\begin{itemize}[leftmargin=*]
    \item Context: It would be perfect to have a family member involved in choosing foster care.
    \item Evidence: Usually children are taken care of by their parents, legal guardians or siblings.
    \item Response: Very true, that is why I think it is best when parents or or legal guardians take care of their children, because they are they only ones that love the children.
    \item Explanation: Although part of the response is present in the evidence, this is a subjective opinion of the speaker.
\end{itemize}

To start the final task, we ask workers to read the dialogue, the corresponding responses, and the Wikipedia knowledge provided (links and pieces of evidence). 
\begin{itemize}[leftmargin=*]
    \item For each provided response, mark them as SUPPORTED, REFUTED, or NOT ENOUGH INFORMATION.
    \item if the response consists of only personal opinions or personal information with no verifiable factual information, please mark the corresponding checkbox.
    \item Please read the instructions and examples in the link above carefully.
    \item If you select the SUPPORTED or REFUTED option, you must click at least one checkbox as evidence or copy-and-paste sentences from Wikipedia links.
    \item For NEI, you would generally need to verify the facts in the responses by visiting and searching Wikipedia pages and pasting any related evidence.
    \item Please edit and correct the responses if they contain any grammatical or spelling mistakes.
\end{itemize}

\end{document}